\documentclass[journal, 12pt]{IEEEtran}
\onecolumn
\usepackage{setspace}
\usepackage{amsmath}
\usepackage{amsthm}   % Provides theorem environments
\usepackage{amssymb} % For additional math symbols
\usepackage[bottom,hang,flushmargin]{footmisc}
\usepackage{graphicx}
\usepackage{caption}
\usepackage{hyperref}
\renewcommand{\footnoterule}{%
  \kern -3pt
  \hrule width 0.4\textwidth height 0.4pt
  \kern 2.6pt
}

\begin{document}

\title{Wav-KAN: Wavelet Kolmogorov-Arnold Networks}

\author{Zavareh Bozorgasl,~\IEEEmembership{Member,~IEEE}, Hao~Chen,~\IEEEmembership{Member,~IEEE}  \thanks{Z. Bozorgasl (zavarehbozorgasl@boisestate.edu) and H. Chen (haochen@boisestate.edu) are with the Department of Electrical and Computer Engineering, Boise State University, Boise,
ID, 83712.} }

% make the title area
\maketitle

\begin{abstract}
In this paper \footnote{Another title for the paper can be "Wavelet for Everybody".}, we introduce Wav-KAN \footnote{Wav-KAN refers to a family of continuous and discrete wavelet transforms in KAN.}, an innovative neural network architecture that leverages the Wavelet Kolmogorov-Arnold Networks (Wav-KAN) framework to enhance interpretability and performance. Traditional multilayer perceptrons (MLPs) and even recent advancements like Spl-KAN \cite{kan} face challenges related to interpretability, training speed, robustness, computational efficiency, and performance. Wav-KAN addresses these limitations by incorporating wavelet functions into the Kolmogorov-Arnold network structure, enabling the network to capture both high-frequency and low-frequency components of the input data efficiently. Wavelet-based approximations employ orthogonal or semi-orthogonal basis and also maintains a balance between accurately representing the underlying data structure and avoiding overfitting to the noise. While continuous wavelet transform (CWT) has a lot of potentials, we also employed discrete wavelet transform (CWT) for multiresolution analysis which obviated the need for recalculation of the previous steps in finding the details \footnote{We demonstrate its efficacy in capturing detailed signal characteristic.} and efficiently combines local detailed information where the data points are dense with broader trends where the data points are sparse. Analogous to how water conforms to the shape of its container, Wav-KAN adapts to the data structure, resulting in enhanced accuracy, faster training speeds, and increased robustness compared to Spl-KAN and MLPs. Our results highlight the potential of Wav-KAN as a powerful tool for developing interpretable and high-performance neural networks, with applications spanning various fields. This work sets the stage for further exploration and implementation of Wav-KAN in frameworks such as PyTorch, TensorFlow, and also it makes wavelet in KAN in wide-spread usage like nowadays activation functions like ReLU and sigmoid in universal approximation theory (UAT). The codes to replicate the simulations are available at \href{https://github.com/zavareh1/Wav-KAN}{https://github.com/zavareh1/Wav-KAN}.  
\end{abstract}


\IEEEpeerreviewmaketitle

\section{Introduction}
Advancements in artificial intelligence (AI) have led to the creation of highly proficient AI systems that make decisions for reasons that are not clear to us. This has raised concerns about the widespread deployment of untrustworthy AI systems in the economy and our daily lives, introducing several new risks, including the potential for future AIs to deceive humans to achieve undesirable objectives \cite{ai_risk_1, ai_risk_2}. The tendency to unravel the black-box behaviour of neural network has attracted a lot attention in recent years. \\
Interpretability of neural networks is crucial as it influences trust in these systems and helps address ethical concerns such as algorithmic discrimination. It is also essential for applying neural networks in scientific fields like drug discovery and genomics, where understanding the model's decisions is necessary for validation and regulatory compliance \cite{surveyInterpret,Interpret_ref}.\\
The multilayer feedforward perceptron (MLP) model is among the most widely used and practical neural network models \cite{MLPapproximation}. Despite their wide-spread usage, MLPs have serious drwabacks like consuming almost all
non-embedding parameters in transformers and also have less interpretability relative to attention layers \cite{attention,kan,autoencoders}.\\

Model renovation approaches aim to enhance interpretability by incorporating more understandable components into a network. These components can include neurons with specially designed activation functions, additional layers with specific functionalities, modular architectures, and similar features \cite{interpret_survey_transition_to_LAN}.\\

Breaking down a neural network into individual neurons has a key challenge of polysemantic neurons, i.e., each neuron activates for several unrelated types of feature \cite{polysemantic}. \cite{basis1} explores the reasons behind the emergence of polysemanticity and propose that it occurs because of superposition, i.e., models learn more distinct features than the available dimensions in a layer. Due to the limitation that a vector space can only contain as many orthogonal vectors as it has dimensions, the network ends up learning an overcomplete set of non-orthogonal features. For superposition to be beneficial, these features must activate sparsely; otherwise, the interference between non-orthogonal features would negate any performance improvements \cite{autoencoders}.\\

\begin{figure}[ht]
    \centering
    \includegraphics[width=0.8\linewidth]{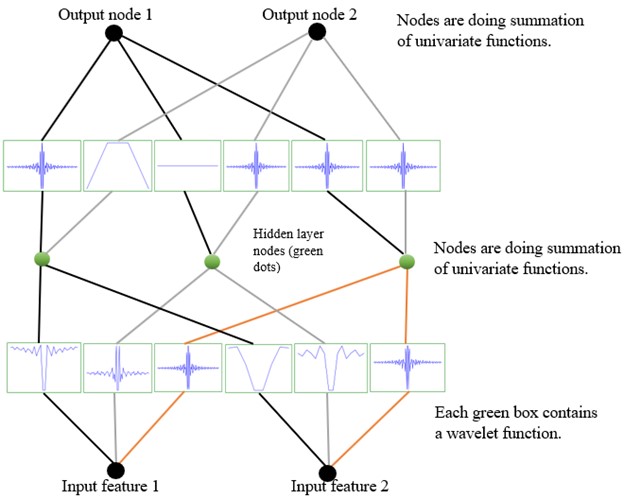} % Ensure the path is correct
    \captionsetup{justification=centering} % Ensure caption is centered
    \caption{Wav-KAN with arbitrary number of layers (here is Wav-KAN[2,3,2])} % Add a caption for the figure
    \label{fig:fig1} % Label should be after the caption
\end{figure}

Kolmogorov-Arnold Networks (KANs) which stand on the Kolmogorov-Arnold representation theorem \cite{kan_old} can bring a lot of advantages including more interpretability and accuracy \cite{kan}. KANs have univariate learnable activation function on edges and nodes are summing those activation functions. The integration of KAN, ensembles of probabilistic trees, and multivariate B-spline representations is presented in \cite{exsplinet}. However, most of the previous works on KAN is for depth-2 representation; except, Liu et. al. \cite{kan} which extends KANs to to arbitrary widths and depths. As \cite{kan} is based on B-Spline, we call it Spl-KAN.\\ 
In this paper, we present an improved version of KAN, called Wav-KAN which uses wavelet in KAN configuration. Figure \ref{fig:fig1} shows a Wav-KAN \footnote{These fancy plots are just to show the potential of wavelet for approximation; they are not output of activation functions of a real network; hence, we see some of them are not differentiable.} with 2 input features, 3 hidden nodes, and 2 output nodes (Wav-KAN[2,3,2]). In general, it can have an arbitrary number of layers. The structure is similar to MLPs with weights replaced by wavelet functions, and nodes are doing summation of those Wavelet functions.

Wavelet has been extensively employed in multiresolution analysis \cite{mallat}. There are some studies of wavelet in neural networks in Universal Approximation Therory (UAT). For example, the complex Gabor wavelet activation function is renowned for its optimal concentration in the space-frequency domain and its exceptional ability to represent images accurately. Hence, \cite{wire} used Gabor wavelet in implicit neural representation (INR) in application including image denoising, image inpainting, super-resolution, computed tomography reconstruction, image overfitting, and novel view synthesis with neural radiance fields. To the best of our knowledge, our proposed framework is the first work which uses wavelet in Kolmogorov-Arnold representation theorem for an arbitrary widths and depth neural netwroks. \\
 In comparison with the recent proposed Spl-KAN \cite{kan} and also MLPs, the proposed configuration is faster, more accurate, and more robust, effectively addressing the existing issues and significantly enhancing its performance.\\
 Also, in statistical learning, we look for models which are more flexible and more interpretable \cite{tibshirani}. Wav-KAN will be a powerful member of a family of KAN glass-box models which facilitates the interpretability of neural networks. This work introduces Wav-KAN and we believe with the potential of Wav-KAN which combines the potential of wavelet and KAN, it will be widely used in all the fields and it will be implemented in Pytorch, Tensor-flow, R, to name but few.\   

Incorporating Wav-KAN in neural networks makes increasingly explainable components that can achieve similar state-of-the-art performance across diverse tasks \footnote{Some of the proofs and experiments will be presented in the final version of the paper.}. \\

Section \ref{sec:KAN} discusses KAN and its generalization to multi-layers KAN  with wavelet as function approximation (i.e., activation function). Section \ref{sec:CWT} presents Continuous Wavelet Transform, especially the definition and criteria for being a mother wavelet which can be used as basis for function approximation. The comparison of Wav-KAN, Spl-KAN and MLPs is given in \ref{sec:comparision}. Some experiment results will be given in Section \ref{sec:sim}. Finally, Section \ref{sec:conclusion} concludes the paper.

%\hfill mds
\section{Kolmogorov-Arnold Networks}
\label{sec:KAN}

Kolmogorov-Arnold Networks (KANs) represent a novel twist on neural network design that challenges traditional concepts like the Multi-Layer Perceptron (MLP). At the heart of KANs is a beautiful and somewhat abstract mathematical theorem from Kolmogorov and Arnold.

\subsection{The Kolmogorov-Arnold Representation Theorem}

Let's start with the theorem that inspires KANs:

\textbf{Theorem: Kolmogorov-Arnold Representation} \cite{kan_old}
    For any continuous function \( f \) of \( n \) variables defined on a cube \( [0,1]^n \), there exist \( 2n+1 \) functions \( \phi_q \) and \( 2n+1 \times n \) functions \( \psi_{q,p} \), all univariate and continuous, such that
    \begin{equation}
        f(x_1, \ldots, x_n) = \sum_{q=1}^{2n+1} \Phi_q \left( \sum_{p=1}^n \phi_{q,p}(x_p) \right).
    \end{equation}

This theorem tells us that any multivariate function can essentially be decomposed into the sum of functions of sums. The magic here is that these inner functions are univariate, meaning they each take a single input. As we use mother wavelet as our basis, in Wav-KAn notation, we use $\psi_{i,j}(x_j)$ instead of $\phi_{i,j}(x_j)$ and $\Psi_i$ instead of $\Phi_i$.

\subsection{From Theory to Networks}

How do we translate this theorem into a neural network architecture? Imagine this: Instead of weights and biases adjusting linear combinations of inputs at each node, KANs modify this process to work with functions. One generalized
version of KAN theorem that corresponds to deeper KANs is the recently published work of Spl-KAN\cite{kan}.  

\begin{itemize}
    \item In KANs, every "weight" is actually a small function on its own. Each node in a KAN does not apply a fixed non-linear activation function. Each learnable activation function in edges, gets the input and gives an output.   
\end{itemize}

Suppose we have an MLPs neural network with \( n \) input and \( m \) output in a fully connected layer (between layer \( l \) and \( l+1 \)). The equation in matrix form is given by:

\begin{equation}
    \mathbf{x}^{(l+1)} = \mathbf{W_{\mathit{l+1,l}}} \mathbf{x}^{(l)} + \mathbf{b}^{(l+1)}
\end{equation}

where:
\begin{itemize}
    \item \( \mathbf{W_{\mathit{l+1,l}}} \) is the weight matrix connecting layer \( l \) and layer \( l+1 \),
    \item \( \mathbf{x}^{(l)} \) is the input vector,
    \item \( \mathbf{x}^{(l+1)} \) is the output vector,
    \item \( \mathbf{b}^{(l+1)} \) is the bias vector for layer \( l+1 \).
\end{itemize}

The weight matrix \( \mathbf{W_{\mathit{l+1,l}}} \) is expanded as follows:

\begin{equation}
\mathbf{W_{\mathit{l+1,l}}} = \begin{pmatrix}
w_{1,1} & w_{1,2} & \cdots & w_{1,n} \\
w_{2,1} & w_{2,2} & \cdots & w_{2,n} \\
\vdots & \vdots & \ddots & \vdots \\
w_{m,1} & w_{m,2} & \cdots & w_{m,n}
\end{pmatrix}
\end{equation}

where $w_{i,j}$, $i=1,2,...,m$ and $j = 1,2, ..., n$ is the weight between $i$-th node in $l+1$-th layer, and $j$-th node in $l$-th .The bias vector \( \mathbf{b}^{(l+1)} \) is:

\begin{equation}
\mathbf{b}^{(l+1)} = \begin{pmatrix}
b_1^{(l+1)} \\
b_2^{(l+1)} \\
\vdots \\
b_m^{(l+1)}
\end{pmatrix}
\end{equation}

Thus, the complete equation becomes:

\begin{equation}
\begin{pmatrix}
x_1^{(l+1)} \\
x_2^{(l+1)} \\
\vdots \\
x_m^{(l+1)}
\end{pmatrix}
=
\begin{pmatrix}
w_{1,1} & w_{1,2} & \cdots & w_{1,n} \\
w_{2,1} & w_{2,2} & \cdots & w_{2,n} \\
\vdots & \vdots & \ddots & \vdots \\
w_{m,1} & w_{m,2} & \cdots & w_{m,n}
\end{pmatrix}
\begin{pmatrix}
x_1^{(l)} \\
x_2^{(l)} \\
\vdots \\
x_n^{(l)}
\end{pmatrix}
+
\begin{pmatrix}
b_1^{(l+1)} \\
b_2^{(l+1)} \\
\vdots \\
b_m^{(l+1)}
\end{pmatrix}
\end{equation}

Now, suppose we have \( L \) layers, each of them having the structure described above. Let \( \sigma(\cdot) \) be the activation function. The compact formula for the whole network, \( f(\mathbf{x}) \), where \( \mathbf{x} \) is the input vector and \( f(\cdot) \) is the neural network, is given by:

\begin{equation}
f(\mathbf{x}) = \mathbf{x}^{(L)}
\end{equation}

where

\begin{equation}
\mathbf{x}^{(l+1)} = \sigma(\mathbf{W_{\mathit{l+1,l}}} \mathbf{x}^{(l)} + \mathbf{b}^{(l+1)})
\end{equation}

for \( l = 0, 1, 2, \ldots, L-1 \), and \( \mathbf{x}^{(0)} = \mathbf{x} \) is the input vector. 
\begin{equation}
f(\mathbf{x}) = \sigma\left(\mathbf{W_L} \sigma\left(\mathbf{W_{L-1}} \cdots \sigma\left(\mathbf{W_2} \sigma\left(\mathbf{W_1} \mathbf{x} + \mathbf{b_1}\right) + \mathbf{b_2}\right) \cdots + \mathbf{b_{L-1}}\right) + \mathbf{b_L}\right)
\end{equation}

In KAN, the relationship between layers turns into:
Let \(\mathbf{x}^{(l)}\) be a vector of size \(n\). We transpose \(\mathbf{x}^{(l)}\) and place it in a matrix \(\mathbf{X}\) with \(m\) rows and \(n\) columns:

\[
\mathbf{x}^{(l)} \in \mathbb{R}^n \quad \Rightarrow \quad (\mathbf{x}^{(l)})^T \in \mathbb{R}^{1 \times n}
\]

We construct the matrix \(\mathbf{X}_l\) as follows:

\[
\mathbf{X}_l = \begin{pmatrix}
(\mathbf{x}^{(l)})^T \\
(\mathbf{x}^{(l)})^T \\
\vdots \\
(\mathbf{x}^{(l)})^T
\end{pmatrix} \in \mathbb{R}^{m \times n}
\]

where each row of \(\mathbf{X}_l\) is the transposed vector \((\mathbf{x}^{(l)})^T\).

We define the operator \( T_o \) which acts on the matrix \(\Psi_{l+1,l} (\mathbf{X}_l)\). This operator sums the elements of each row of the matrix and outputs the resulting vector \(\mathbf{v}\). The definition is as follows:

\[
T_o\left(\Psi_{l+1,l} (\mathbf{X}_l)\right) = \mathbf{v}
\]

where \(\mathbf{v}\) is a vector with elements given by:

\[
v_i = \sum_j \left[ \Psi_{l+1,l} (\mathbf{X}_l) \right]_{ij} =\sum_{j=1}^{n} \psi_{i,j}(x_j^{(l)}), \quad \text{for } i = 1, 2, \ldots, m
\]

In this expression, \( \left[ \Psi_{l+1,l} (\mathbf{X}_l) \right]_{ij} \) represents the element in the \(i\)-th row and \(j\)-th column of the matrix \(\Psi_{l+1,l} (\mathbf{x}^{(l)})\).

Thus, the operator \( T_o \) can be written as:

\[
T_o\left(\Psi_{l+1,l} (\mathbf{X}_l)\right) = \left( \sum_j \left[ \Psi_{l+1,l} (\mathbf{X}_l) \right]_{ij} \right)_{i}
\]

In this definition, \( T_o \) takes the matrix \(\Psi_{l+1,l} (\mathbf{X}_l)\), sums the elements of each row, and outputs the resulting vector \(\mathbf{v}\).

Indeed \(\Psi_{l+1,l}\) acts on the input vector \(\mathbf{x}^{(l)}\) and gives an output where each element of \(\Psi_{l+1,l}\) takes one corresponding element of \(\mathbf{x}^{(l)}\) as input, sums up the results, and produces one element of the output:

\begin{equation}
\label{eq:kan_rel_layers}
   \mathbf{X}_{l+1} = \Psi_{l+1,l} (\mathbf{X}_{l}) 
\end{equation}

where:

\begin{equation}
\Psi_{l+1,l} (\mathbf{X}_{l})  = \begin{pmatrix}
\psi_{1,1}(x_1^{(l)}) & \psi_{1,2}(x_2^{(l)}) & \cdots & \psi_{1,n}(x_n^{(l)}) \\
\psi_{2,1}(x_1^{(l)}) & \psi_{2,2}(x_2^{(l)}) & \cdots & \psi_{2,n}(x_n^{(l)}) \\
\vdots & \vdots & \ddots & \vdots \\
\psi_{m,1}(x_1^{(l)}) & \psi_{m,2}(x_2^{(l)}) & \cdots & \psi_{m,n}(x_n^{(l)})
\end{pmatrix}
\end{equation}

Here, \( \Psi_{l+1,l} \) represents the activation functions connecting layer \( l \) and layer \( l+1 \). Each element \( \psi_{i,j}(\cdot) \) denotes the activation function that connects the \( j \)-th neuron in layer \( l \) to the \( i \)-th neuron in layer \( l+1 \). Instead of multiplication, equation (\ref{eq:kan_rel_layers}) computes a function of input with distinct learnable parameters.

Hence, if $\mathbf{X}_{0}$ be considered as input which just contains input vector as its rows, for the entire network, the output after \(L\) layers is:

\[
f_{KAN}(\mathbf{X}_{0})=\mathbf{x}^{(L)} = T_o\left(\Psi_{L,L-1} \left(\mathbf{X}_{L-1} \right)\right) = T_o\left(\Psi_{L,L-1} \left( \begin{pmatrix}
(T_o(\Psi_{L-1,L-2} (\mathbf{X}_{L-2})))^T \\
(T_o(\Psi_{L-1,L-2} (\mathbf{X}_{L-2})))^T \\
\vdots \\
(T_o(\Psi_{L-1,L-2} (\mathbf{X}_{L-2})))^T
\end{pmatrix} \right)\right)
\]

\begin{equation}
   = \cdots = T_o\left(\Psi_{L,L-1} \left(\begin{pmatrix}
(T_o(\Psi_{L-1,L-2} \cdots (T_o(\Psi_{1,0} (\mathbf{X}_0))))^T \\
(T_o(\Phi_{L-1,L-2} \cdots (T_o(\Psi_{1,0} (\mathbf{X}_0))))^T \\
\vdots \\
(T_o(\Psi_{L-1,L-2} \cdots (T_o(\Psi_{1,0} (\mathbf{X}_0))))^T
\end{pmatrix} \right)\right) 
\end{equation}

In summary, traditional MLPs use fixed nonlinear activation functions at each node and linear weights (and biases) to transform inputs through layers. The output at each layer is computed by a linear transformation followed by a fixed activation function. During backpropagation, gradients of the loss function with respect to weights and biases are calculated to update the model parameters. In contrast, KANs replace linear weights with learnable univariate functions placed on edges rather than nodes. In the nodes, we just have a summation of some univariate functions from previous layers. Each function is adaptable, allowing the network to learn both the activation and transformation of the inputs. This change leads to improved accuracy and interpretability, as KANs can better approximate functions with fewer parameters. During backpropagation in KANs, the gradients are computed with respect to these univariate functions, updating them to minimize the loss function. This results in more efficient learning for complex and high-dimensional functions.

\subsection{Why Bother with KANs?}

The flexibility of KANs allows for a more nuanced understanding and adaptation to data. By learning the functions directly involved in data relationships, KANs aim to provide a more accurate and interpretable model:
\begin{itemize}
    \item \textbf{Accuracy:} They can fit complex patterns in data more precisely with potentially fewer parameters.
    \item \textbf{Interpretability:} Since each function has a specific, understandable role, it's easier to see what the model is "thinking."
\end{itemize}

In summary, KANs leverage deep mathematical insights to offer a fresh perspective on how neural networks can understand and interact with the world. By focusing on functions rather than mere weights, they promise a richer and more intuitive form of machine learning.

\section{Continuous Wavelet Transform}
\label{sec:CWT}

The Continuous Wavelet Transform (CWT) is a method mostly used in signal processing to analyze the frequency content of a signal as it varies over time \cite{mallat}. It acts like a microscope, zooming in on different parts of a signal to determine its constituent frequencies and their variations.

CWT utilizes a base function known as a ``mother wavelet,'' which serves as a template that can be scaled and shifted to match various parts of the signal. The shape of the mother wavelet is critical as it dictates which features of the signal are highlighted.

Let $\psi \in L^2(\mathbb{R})$ be the mother wavelet and $g(t) \in L^2(\mathbb{R})$\footnote{\[L^2(\mathbb{R}) = \left\{ f(x) \mid \int |f(x)|^2 \, dx < \infty \right\} \].} be the function that we want to express in the wavelet basis. Then, a mother wavelet must satisfy certain criteria \cite{calderon1964,grossmann1984}.

\begin{enumerate}
  \item \textbf{Zero Mean:} The integral of the wavelet over its entire range must equal zero:\
  \begin{equation}
    \int_{-\infty}^{\infty} \psi(t) \, dt = 0  
  \end{equation}

  \item \textbf{Admissibility Condition:} The wavelet must have finite energy, which means the integral of the square of the wavelet must be finite. 
\begin{equation}
    C_\psi = \int_{0}^{+\infty} \frac{|\hat{\psi}(\omega)|^2}{\omega} \, d\omega < +\infty
\end{equation}

where $\hat{\psi}$ is the Fourier transform of the wavelet $\psi(t)$.
\end{enumerate}

The CWT of a signal/function is represented by wavelet coefficients, calculated as follows:
\begin{equation}
C(s, \tau) = \int_{-\infty}^{+\infty} g(t) \frac{1}{\sqrt{s}} \psi\left(\frac{t-\tau}{s}\right) dt
\end{equation}

where:
\begin{itemize}
  \item \(g(t)\) is the signal/function that we want to approximate by Wavelet basis.
  \item \(\psi(t)\) is the mother wavelet.
  \item \( s \in \mathbb{R}^{+} \) is the scale factor which is greater than zero.
  \item \( \tau \in \mathbb{R} \) is the shift factor.
  \item \(C(s, \tau)\) measures the match between the wavelet and the signal at scale \(s\) and shift \(\tau\).
\end{itemize}

A signal can be reconstructed from its wavelet coefficients using the inverse CWT:
\begin{equation}
g(t) = \frac{1}{C_\psi}  \int_{-\infty}^{+\infty} \int_{0}^{+\infty} C(s, \tau) \frac{1}{\sqrt{s}} \psi\left(\frac{t-\tau}{s}\right) \frac{ds\,d\tau}{s^2}
\end{equation}

where \(C_\psi\) is a constant that depends on the wavelet, ensuring the reconstruction's accuracy.

\section{Discrete Wavelet Transform}
\label{sec:DWT}

The Discrete Wavelet Transform (DWT) is a widely used method in signal processing that decomposes a signal into different frequency components with high efficiency \cite{mallat}. Unlike the Continuous Wavelet Transform (CWT), which analyzes the signal at every possible scale and translation, the DWT uses discrete sampling intervals which is suitable for digital applications. When the data points are close together (high sampling rate), the wavelet basis can zoom in to capture the high-frequency details (fine details) in that local area. When the data points are spread out (low sampling rate), the wavelet basis can zoom out to capture the low-frequency trends (overall shape) using global information. We see some examples uniform and irregular samples of data. Indeed, by employing wavelet we combine local detailed information where the data points are dense with broader trends where the data points are sparse.

The DWT employs a set of base functions derived from a single mother wavelet through scaling and translation. These base functions are orthonormal, ensuring that the DWT provides a compact and non-redundant representation of the signal. By applying high-pass and low-pass filters iteratively to the signal, the DWT produces approximation and detail coefficients at various levels of resolution, allowing for a multi-resolution analysis.

Let $\psi \in L^2(\mathbb{R})$ be the mother wavelet and $\phi \in L^2(\mathbb{R})$ be the scaling function. The signal $g(t) \in L^2(\mathbb{R})$\footnote{\[L^2(\mathbb{R}) = \left\{ f(x) \mid \int |f(x)|^2 \, dx < \infty \right\} \].} is decomposed into approximation coefficients $a_j(k)$ and detail coefficients $d_j(k)$ at level $j$ as follows:

\begin{equation}
a_j(k) = \sum_{n} g(n) \phi_{j,k}(n)
\end{equation}

\begin{equation}
d_j(k) = \sum_{n} g(n) \psi_{j,k}(n)
\end{equation}

where:
\begin{itemize}
  \item $g(n)$ is the discrete signal.
  \item $\phi_{j,k}(n)$ is the scaling function at scale $j$ and position $k$.
  \item $\psi_{j,k}(n)$ is the wavelet function at scale $j$ and position $k$.
  \item $a_j(k)$ represents the approximation coefficients.
  \item $d_j(k)$ represents the detail coefficients.
\end{itemize}

The original signal can be reconstructed from its wavelet coefficients using the inverse DWT, ensuring that no information is lost during the transformation process. This reconstruction is given by:

\begin{equation}
g(n) = \sum_{k} a_j(k) \phi_{j,k}(n) + \sum_{j=1}^{J} \sum_{k} d_j(k) \psi_{j,k}(n)
\end{equation}

where $J$ is the number of decomposition levels.

The DWT's ability to provide both time and frequency localization, coupled with its computational efficiency, makes it an indispensable tool in various fields such as image compression, noise reduction, and feature extraction.

Multi-resolution analysis (MRA) using wavelets, specifically discrete wavelet transforms (DWT), is a powerful technique in signal processing and data analysis. MRA allows the decomposition of a signal into different levels of detail by using wavelets, which are localized functions. The DWT achieves this by recursively applying high-pass and low-pass filters to the data/signal, creating approximations and details at various resolutions. This process provides a hierarchical framework, where each level captures different frequency components of the signal, making it easier to analyze localized features and transient phenomena. The ability to zoom in on fine details while retaining the broader structure of the signal is a key advantage of MRA, making it useful in applications such as image compression, noise reduction, and feature extraction.

In the context of discrete wavelet transforms, the data/signal is represented as a sum of wavelet functions at different scales and positions. In wavelet analysis, once we compute the initial set of coefficients, we do not need to recompute them when calculating additional coefficients at finer resolutions. This efficiency arises because wavelet transforms leverage previously computed coefficients to generate further details. Consequently, the hierarchical nature of the multi-resolution analysis ensures that the earlier computations remain valid and reusable, streamlining the process and significantly reducing the computational overhead. This property makes wavelet-based methods particularly advantageous for iterative and real-time signal processing applications. The multi-resolution property of wavelets is particularly beneficial for analyzing non-stationary signals, which have frequency components that vary over time. By breaking down the signal into different resolution levels, MRA enables a more comprehensive understanding of its structure and characteristics. Moreover, the DWT's ability to provide both time and frequency localization offers a significant advantage over traditional Fourier transforms, which only provide frequency information \footnote{We will add some simulations to this section.}.

In summary, wav-kan by employing wavelet adapts to the varying density of data points, using more detailed information where there are more data points and less detailed information where there are fewer data points. This approach is particularly useful for real-world data that often suffers from irregular sampling and data dropouts.

\section{Wav-KAN or Spl-KAN or MLPs?}
\label{sec:comparision}
Wavelets and B-splines are two prominent methods used for function approximation, each with distinct advantages and limitations, particularly when applied in neural networks. B-splines provide smooth and flexible function approximations through piecewise polynomial functions defined over control points. They offer local control, meaning adjustments to a control point affect only a specific region, which is advantageous for precise function tuning. This smoothness and local adaptability make B-splines suitable for applications requiring continuous and refined approximations, such as in CAD and computer graphics. However, the computational complexity increases significantly with higher dimensions, making them less practical for high-dimensional data. Managing knot placement can also be intricate and affect the overall shape and accuracy of the approximation. While B-splines can be used in neural networks to approximate activation functions or smooth decision boundaries, their application is generally less suited to feature extraction tasks compared to wavelets due to their limited ability to handle multi-resolution analysis and sparse representations.\\

Wavelets excel in multi-resolution analysis  enabling different levels of detail to be represented simultaneously which makes it a precious tool to decompose data into various frequency components. This capability is highly beneficial for feature extraction in neural networks, as it enables capturing of both high-frequency details and low-frequency trends. Additionally, wavelets offer sparse representations, which can lead to more efficient neural network architectures and faster training times. They are also well-suited for handling non-stationary signals and localized features, making them ideal for applications such as image recognition and signal classification. However, the choice of the wavelet function is crucial and can significantly impact performance, and edge effects can introduce artifacts that need special handling.\\

In function approximation, wavelets excel by maintaining a delicate balance between accurately representing the underlying data structure and avoiding overfitting to the noise. Unlike traditional methods that may overly smooth the data or fit to noise, wavelets achieve this balance through their inherent ability to capture both local and global features. By decomposing the data into different frequency components, wavelets can isolate and retain significant patterns while discarding irrelevant noise. This multiresolution analysis ensures that the approximation is robust and reliable, providing a more accurate and nuanced representation of the original data without the pitfalls of noise overfitting. On the other side, while Spl-KAN is powerful in capturing the changes in data, it also captures the noise in training data. Indeed, the strength of Spl-KAN is its weakness, too.\\
The advantages of Spl-KAN which are mentioned in \cite{kan} including interpretability and/or accuracy with respect to MLPs exist for Wav-KAN. More importantly Wav-KAN has solved the major disadvantage of Spl-KANs which was slow training speed. In terms of number of parameter, we compared Wav-KAN with Spl-KAN and MLPs for a hypothetical neural network which has $N$ input nodes and $N$ output node, with $L$ layers. As we see in Table \ref{tab:spl_wav_mlp_order} and by considering the value of G, Wav-KAN has less number of parameters than Spl-KAN (k should be at least 2 to have good results in Spl-KAN, especially in complex tasks). The coefficient 3 is becuase Wav-KAN has a learnable weight, a translation and a scaling. Learnable parameters in each neural network is in column of parameters. While the order of MLPs are less for the hypothetical neural network, in practice, Wav-KAN needs less number of parameters to learn the same task. Indeed, this originates because of capacity of Wavelet for capturing both low frequency and high frequency functions.

\begin{table}[h!]
\centering
\caption{Comparison of MLPs, Spl-KAN and Wav-KAN}
\label{tab:spl_wav_mlp_order}
\begin{tabular}{|c|c|c|}
\hline
\multicolumn{3}{|c|}{Neural network With $L$ layers, each layer has $N$ nodes} \\
\hline
Neural Network Structure & Order & Parameters \\
\hline
MLPs & $O(N^2L)$ or $O(N^2L+NL)$ & weights and biases \\
\hline
Spl-KAN & $O(N^2L(G + k + 1)) \sim O(N^2LG)$ & weights \\
\hline
Wav-KAN & $O(3N^2L)$  & weight, translation, scaling \\
\hline
\end{tabular}

\end{table}

Regarding the implementation, Spl-KAN requires a smooth function like $b(x)$ in the equation (2.10) \cite{kan} attempting to catch on some global features. Because of the inherent scaling property of Wavelet, Wav-KAN does not need an additional term in its activation functions. This helps Wav-KAN to be faster. For example, 
wavelets equal to the second derivative of a Gaussian  which are called \textit{Mexican hats} \footnote{The following equation is normalized Mexican hat wavelet. Indeed, pywavelets has a minus sign behind it which because of the weight we put behind the mother wavelet in our activation functinos, the sign  doesn't matter(https://pywavelets.readthedocs.io/en/latest/ref/cwt.html).} and first used in computer vision to detect multiscale edges \cite{witkin,mallat} has the following form 

\begin{equation}
    \psi(t) = \frac{2}{\pi^{1/4} \sqrt{3\sigma}} \left( \frac{t^2}{\sigma^2} - 1 \right) \exp \left( \frac{-t^2}{2\sigma^2} \right). 
\end{equation}

Where $\sigma$ shows the adjustable standard deviation of Gaussian. In our experiments, $\psi_{exp}(t)$ 
\begin{equation}
  \psi_{exp}(t) = w \psi(t)  
\end{equation}
 Indeed, $w$ plays the role of CWT coefficients which is multiplied by the mother wavelet formula; as $w$ is a learnable parameter, it helps adapting the shape of the mother wavelet to the function that it tries to approximate .

 Moreover, Spl-KAN heavily depends on the grid spaces, and for better performance, it requires increasing the number of grids \footnote{Which is equivalent to decrease the spaces between the grids. This helps Spl-KAN accuracy to be improved.}, though, it brings two disadvantages. First it needs curvefitting which is a cumbersome and computational expensive operation, and also while we increase the number of grids, loss has some jumps \footnote{Loss will be increased for some training, then, it works better. See Figure 2.3 \cite{kan} for better illustration.}. Fortunately, wavelet is safe from such computations and deficiencies and if one wants to capture more details, DWT efficiently does that without recalculation of the previous steps. \\

 Last but no least, we found that batch normalization \cite{batch} significantly improves accuracy and speeds up training of both Wav-KAN and Spl-KAN; hence, we included batch normalization in both of these methods \footnote{In \cite{kan}, the authors did not mention and didn't apply batch normalization.}.
 
\section{Simulation Results}
\label{sec:sim}
In this section, we present the results of our experiments conducted using the KAN (Kernel-based Artificial Neural network) model with various continuous wavelet transformations (CWTs) \footnote{We add more simulations including some discrete wavelet ones for MRA, etc.}  on the MNIST dataset, utilizing a training set of 60,000 images and a test set of 10,000 images. It is important to note that our objective was not to optimize the parameters to their best possible values, but rather to demonstrate that Wav-KAN performs well in terms of overall performance. We have incorporated batch normalization into both Spl-KAN and Wav-KAN, resulting in improved performance. The wavelet types considered in our study include Mexican hat, Morlet, Derivative of Gaussian (DOG), and Shannon (see Table \ref{tab:mother_wavelets}). For each wavelet type and also Spl-KAN, we performed five trials, training the model for 50 epochs per trial.\\ 

\begin{table}[htbp]
    \centering
    \caption{Mother Wavelet Formulas and Parameters}
    \label{tab:mother_wavelets}
    \begin{tabular}{|c|c|c|}
        \hline
        \textbf{Wavelet Type} & \textbf{Formula of Mother Wavelet} & \textbf{Parameters} \\
        \hline
        Mexican hat & $\psi(t) = \frac{2}{\sqrt{3}\pi^{1/4}} \left(t^2 -1\right) e^{-\frac{t^2}{2}}$ &  $\tau$, $s$  \\
        \hline
       Morlet & $\psi(t) = \cos(\omega_0 t) e^{-\frac{t^2}{2}}$ & $\tau$, $s$ and scaling, $\omega_0$= 5 \\
        \hline
        Derivative of Gaussian (DOG) & $\psi(t) = -\frac{d}{dt}\left( e^{-\frac{t^2}{2}} \right)$ & $\tau$, $s$ \\
        \hline
 
        Shannon & $\psi(t) = \text{sinc}(t/\pi) \cdot w(t)$ & $\tau$, $s$, and $w(t)$: window function \\
        \hline
    \end{tabular}

\end{table}

The results were averaged across these trials to ensure the robustness and reliability of our findings.\\ Both Wav-KAN and Spl-KAN have the structure (number of nodes) of  [28*28,32,10] \footnote{[first layer nodes, middle layer nodes, output nodes]}. Although we enhanced Spl-KAN by using spline order 3 and a grid size of 5, this approach is computationally much more expensive compared to Wav-KAN. We employed AdamW optimizer \cite{adam,adamw}, with learning rate of 0.001 with weight decay of $10^{-4}$. Loss is cross entropy.

\begin{figure}[ht]
    \centering
    \includegraphics[width=0.8\linewidth]{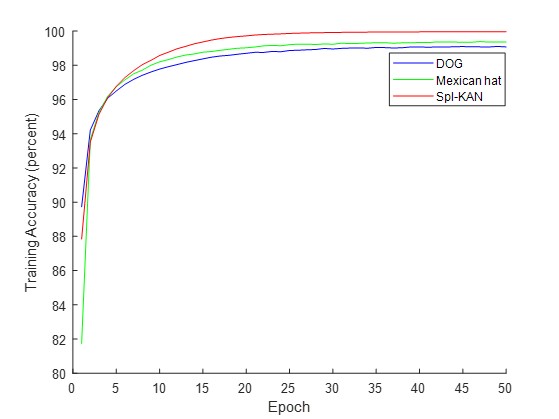} % Ensure the path is correct
    \captionsetup{justification=centering} % Ensure caption is centered
    \caption{ Training accuracy of Wav-KAN [28*28,32,10] versus Spl-KAN [28*28,32,10]} % Add a caption for the figure
    \label{fig:MNIST_train_acc} % Label should be after the caption
\end{figure}

\begin{figure}[ht]
    \centering
    \includegraphics[width=0.8\linewidth]{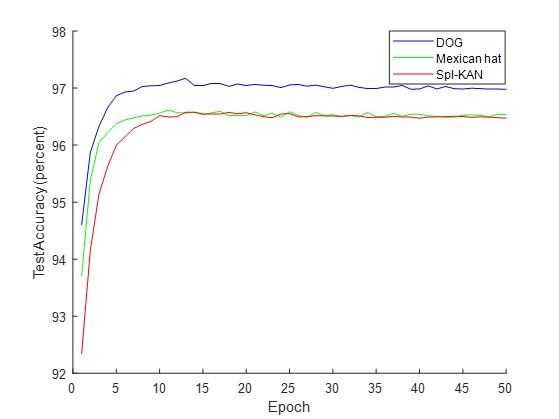} % Ensure the path is correct
    \captionsetup{justification=centering} % Ensure caption is centered
    \caption{ Test accuracy of Wav-KAN [28*28,32,10] versus Spl-KAN [28*28,32,10]} % Add a caption for the figure
    \label{fig:MNIST_test_acc} % Label should be after the caption
\end{figure}

Figures \ref{fig:MNIST_train_acc} and \ref{fig:MNIST_test_acc} show the result of training accuracy and test accuracy of Wav-KAN in comparison to Spl-KAN. To not clutter up, we just show the result of Derivative of Gaussian (DOG) and Mexican hat wavelet. These have been shown as a sample. By fine tuning, and using all the freedom of wavelet (like frequency of sinusoid and variance of Gaussian), Wavelet shows significant superiority \footnote{We have done a lot of such experiments and we will publish them soon.}. While Spl-KAN has better performance in training, which is because of overfitting to data, a lot of wavelet types have shown superior performance with respect to Spl-KAN. Indeed, for this experiment we set the variance of the Gaussian in wavelets to be 1; though, we can find better variances by grid search or making it a learnable parameter. Also, one may like to consider $\omega_0$ in Morlet and parameters of frequency and/or windowing in Shannon wavelet learnable. Indeed, we can use different degrees of freedom that we have to make the shape of wavelets more flexible. Wavelet makes a balance by not fitting to the noise in the data.  

We evaluated the performance of each wavelet type in terms of training loss, training accuracy, validation loss, and validation accuracy. Figures 1 and 2 summarize the results, depicting the training and validation metrics averaged over the five trials for each wavelet type.\\
The simulation results indicate that the choice of wavelet significantly impacts the performance of the KAN model.  This suggests that these wavelets are particularly effective at capturing the essential features of the MNIST dataset while maintaining robustness against noise. On the other hand, wavelets like Shannon and Bump did not perform as well, highlighting the importance of wavelet selection in designing neural networks with wavelet transformations.

\section{Conclusion}
\label{sec:conclusion}

In this paper, we have introduced Wav-KAN, a novel neural network architecture that integrates wavelet functions within the Kolmogorov-Arnold Networks (KAN) framework to enhance both interpretability and performance. By leveraging the multiresolution analysis capabilities of wavelets, Wav-KAN effectively captures complex data patterns and provides a robust solution to the limitations faced by traditional multilayer perceptrons (MLPs) and recently proposed Spl-KANs.

Our experimental results  demonstrate that Wav-KAN not only achieves superior accuracy but also benefits from faster training speeds compared to Spl-KAN. The unique structure of Wav-KAN, which combines the strengths of wavelet transforms and the Kolmogorov-Arnold representation theorem, allows for more efficient parameter usage and improved model interpretability.

Wav-KAN represents a significant advancement in the design of interpretable neural networks. Its ability to handle high-dimensional data and provide clear insights into model behavior makes it a promising tool for a wide range of applications, from scientific research to industrial deployment. Future work will focus on further optimizing the Wav-KAN architecture, exploring its applicability to other datasets and tasks, and implementing the framework in popular machine learning libraries such as PyTorch and TensorFlow.

Overall, Wav-KAN stands out as a powerful and versatile model, paving the way for the development of more transparent and efficient neural network architectures. Its potential to combine high performance with interpretability marks a crucial step forward in the field of artificial intelligence.

\ifCLASSOPTIONcaptionsoff
  \newpage
\fi

\bibliographystyle{IEEEtran}
\bibliography{references}

\end{document}